\documentclass[10pt,twocolumn,letterpaper]{article}

\usepackage[pagenumbers]{cvpr} 

\usepackage{graphicx}
\usepackage{amsmath}
\usepackage{amssymb}
\usepackage{booktabs}

\usepackage{enumitem}
\setitemize{noitemsep,topsep=0pt,parsep=0pt,partopsep=0pt}

\usepackage[pagebackref,breaklinks,colorlinks]{hyperref}

\usepackage[capitalize]{cleveref}
\crefname{section}{Sec.}{Secs.}
\Crefname{section}{Section}{Sections}
\Crefname{table}{Table}{Tables}
\crefname{table}{Tab.}{Tabs.}

\def\cvprPaperID{*****} 
\def\confName{CVPR}
\def\confYear{2023}

\begin{document}

\title{Edge Enhanced Image Style Transfer via Transformers}

\author{
Chiyu Zhang\textsuperscript{1}, Jun Yang\textsuperscript{2}, Zaiyan Dai\textsuperscript{3}, Peng Cao\textsuperscript{4}\\
Sichuan Normal University\\
{\tt\small \textsuperscript{1}alienzhang19961005@gmail.com \textsuperscript{2}jkxy\_yjun@sicnu.edu.cn} \\
{\tt\small \{\textsuperscript{3}daizaiyan, \textsuperscript{4}pc\}@stu.sicnu.edu.cn}
}
\maketitle

\begin{abstract}
   In recent years, arbitrary image style transfer has attracted more and more attention. Given a pair of content and style images, a stylized one is hoped that retains the content from the former while catching style patterns from the latter. However, it is difficult to simultaneously keep well the trade-off between the content details and the style features. To stylize the image with sufficient style patterns, the content details may be damaged and sometimes the objects of images can not be distinguished clearly. For this reason, we present a new transformer-based method named STT for image style transfer and an edge loss which can enhance the content details apparently to avoid generating blurred results for excessive rendering on style features. Qualitative and quantitative experiments demonstrate that STT achieves comparable performance to state-of-the-art image style transfer methods while alleviating the content leak problem.
\end{abstract}

\section{Introduction}
Rendering a content image into the artistic style of a referenced image is the main purpose of the image style transfer. Image style transfer is an interesting topic of computer vision with a long history. About 2 decades ago, researchers \cite{p1,p2} utilize techniques, such as texture synthesis and style transfer functions, to achieve the stylizing process. But they only focus on low-level features. Thereafter Gatys et al. \cite{p3,p4} prove that the correlation between features extracted from a pre-trained VGG can be treated as the representation of style patterns, which opens the gate of neural style transfer (NST). Iterative methods \cite{p4,p5,p6,p7,p8,p9,p10} render the content images gradually by applying gradients on the input images or the noise images while the feed-forward networks \cite{p11,p12,p13,p14,p15,p16,p17,p18,p19} can complete the stylizing process in one feed-forward manner after training. Vivid results though the iterative and feed-forward methods may produce, are still limited to a certain number of styles or achieve inadequate style quality. Thanks to the encoder-transfer-decoder architecture, arbitrary style transfer methods \cite{p20,p21,p22,p23,p24,p25,p26,p27,p28,p29,p30,p31,p32,p33,p34,p35,p36,p37,p38,p39,p40,p41,p42,p43,p44,p45,p46,p89} are capable of rendering images into any styles. However, these models may not work well in some cases due to the limited ability to merge the content and style features. To cope with this problem, the attention mechanism \cite{p47,p48} is introduced by a few methods \cite{p37,p38,p39,p40,p41,p42} to enhance the fusion effects.

\begin{figure}[t]
\centerline{\includegraphics[width=1.0\linewidth]{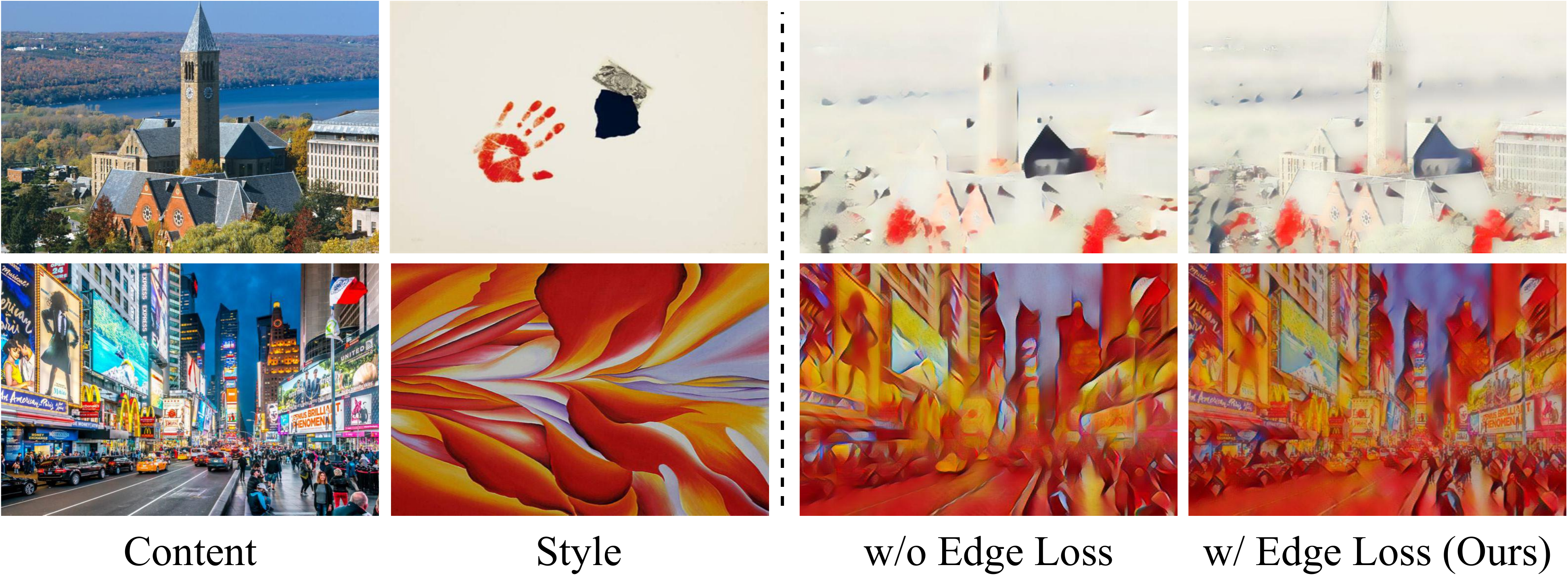}}
\caption{Visual effects of edge loss. Compared to the results from the model without edge loss (column 3), the objects of stylized images with edge loss, especially the small ones like letters or windows, are apparently more clear and distinguishable (column 4). For the convenience of comparison, the model used above is STT. The results from other methods are also blurred in this case.}
\label{fig1}
\end{figure}

Recently, An et al. \cite{p43} discover the content leak problem that the structure of results from the CNN-based methods will be dramatically changed after a few rounds of repetitive stylization process. Deng et al. \cite{p45} and Zhang et al. \cite{p89} then prove that the transformer-based methods are capable of alleviating the problem. Different from the previous methods, IEST \cite{p42} and CAST \cite{p46} leverage the contrastive learning strategy to enhance the visual quality. However, in some cases, the structure of results from previous methods still could be blurred and the objects in images are difficult to be distinguished (see Fig.~\ref{fig1}).

Thanks to the flexibility, scalability, and ability to capture long-range dependencies, Transformers \cite{p48,p49,p50} have been widely used in all kinds of vision tasks. Owing to the self-attention mechanism, Transformer can efficiently gather global information which is important for preserving the structure of input images. Compared to the architecture of typical CNN-based methods, Transformer evades the multi-time downsample operations which may lead to the content leak problem \cite{p43}. Therefore, the Transformer structure has a good effect on the image style transfer task.

In this work, we propose a new Transformer-based image style transfer algorithm that is capable of producing stylized results with high visual quality while preserving fine content details. We call it STT (\textbf{S}tyle \textbf{T}ransfer on \textbf{T}ransformers). Different from StyTr2 \cite{p45} and S2WAT \cite{p89}, neither does STT choose to encode content and style features in different encoders as StyTr2 did, nor does STT adopts the hierarchal structure as S2WAT did. A tiny CNN-based module is equipped as a positional encoding layer (Conv PE) which extracts the positional encoding (PE) according to the semantic information. Furthermore, to enhance the content details, a novel edge loss is applied as an extra restriction when stylized images are blurred due to the overdose of style features imposed on the inputs.

The main contributions of our work are as follows:
\begin{itemize}
\item A new image style transfer network name STT which can stylize images with high quality while preserving fine content details.
\item A novel edge loss to enhance the content details, which improves the picture clarity of the stylized images obviously.
\item Extensive experiments demonstrate that STT achieves outstanding effects and is capable of generating favorable results while preserving fine content details.
\end{itemize}

\section{Related Work}
{\flushleft \bf Image Style Transfer}. Starting from Gatys et al. \cite{p3,p4}, the number of methods in NST is increasingly growing with time forward and the stylizing effects have been more and more colorful. Here we make a rough classification of these models with respect to their generalization abilities, the backbone architecture, and training strategies. In generalization abilities, the categories can be divided into single style transfer \cite{p4,p5,p6,p7,p8,p9,p10,p11,p12,p13,p14,p15}, multiple style transfer \cite{p16,p17,p18}, and arbitrary style transfer \cite{p20,p21,p22,p23,p24,p25,p26,p27,p28,p29,p30,p31,p32,p33,p34,p35,p36,p37,p38,p39,p40,p41,p42,p43,p44,p45,p46,p89}. The single style transfer encodes the fixed style features into models while the multiple style transfer utilizes certain tricks, such as conditional instance normalization \cite{p16} and StyleBank \cite{p17}, to support a number of styles. With a certain module to merge the features of content and style, the arbitrary style transfer is capable of handling any style transfer. As the techniques of upstream tasks like image classification and image generation have been rapidly developing these years, an increasing number of sorceries have been introduced into image style transfer. Except for the typical CNN-based methods, the Flow-based \cite{p43} and the Transformer-based \cite{p45,p89} methods also appear in recent years. The Flow-based ArtFlow \cite{p43} is proposed to solve the problem of the content leak and the Transformer-based StyTr2 \cite{p45} and S2WAT \cite{p89} are able to alleviate the problem. The encoders of StyTr2 have the traditional Transformer structure where the shape of representations will not be changed in processing while S2WAT adopts hierarchal architecture which means the features will be downscaled gradually. Recently, the contrastive learning strategy is introduced by IEST \cite{p42} and CAST \cite{p46}. Different from other methods trained with perceptual losses or identity losses, IEST and CAST treat the contrastive loss and the adversarial loss as the optimization targets to achieve satisfying effects. However, in some cases, the results from the existing image style transfer methods may still be blurred due to no restriction on the edge of the main objects in inputs.

{\flushleft \bf Vison Transformer}. Inherited the ability to capture the long-range dependencies from Transformers in natural language processing (NLP), vision Transformers have been developed in a wide variety of vision tasks, including image classification \cite{p51,p52,p53,p54,p55,p56,p57,p58,p59,p60,p61,p62,p63}, object detection \cite{p67,p68,p69,p70,p71}, semantic segmentation \cite{p72,p73}, and image generation \cite{p75,p76}. In image style transfer, StyTr2 and S2WAT have demonstrated that both the traditional structure and the hierarchical architecture have a favorable effect on style transfer. In this paper, we leverage several convolutional operations to fulfill the positional encoding instead of parametric positional encoding \cite{p51} or the one with pooling operations \cite{p45}.

{\flushleft \bf Utilization of Edge Maps in Style Transfer}. The operators like Laplacian, Canny, and Sobel are widely used in edge and contour detection. In image style transfer, Li et al. discover the correspondences between Laplacian deviations and image distortions and then propose Lapstyle \cite{p8}, an iterative image style transfer method based on a Laplacian loss. Subsequently, Li et al. apply the Laplacian filter on the drafting and revision network and then present LapStyle \cite{p36}, a feed-forward image style transfer method based on the Laplacian filter. However, the above methods are all applied to the CNN-based models. In this work, we leverage the edge maps to enhance the results from the Transformer-based STT which is capable of producing stylized images with fine content details  and colorful artistic features.

\begin{figure*}[htbp]
\centerline{\includegraphics[width=1.0\linewidth]{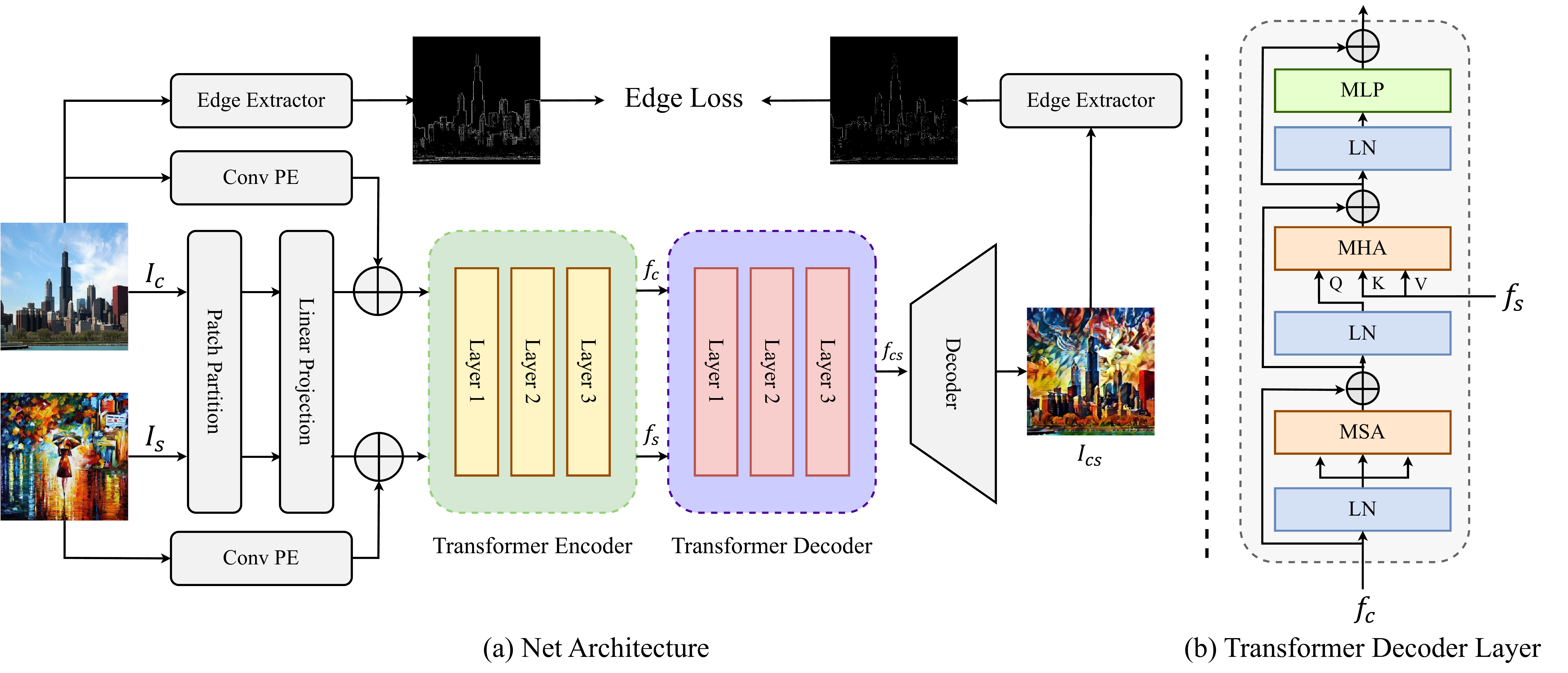}}
\caption{The net architecture of the proposed STT.}
\label{fig2}
\end{figure*}

\section{Method}
As shown in Fig.~\ref{fig2}, the proposed STT has the architecture of encoder-transfer-decoder. The positional encoding (PE) is first extracted from both the content images $ I_c $ and the style image $ I_s $ by a module name Conv PE. Then after splitting the content images $ I_c $ and style image $ I_s $ into non-overlapping patches, a linear projection is equipped to transform the patches into sequences. The sum of the sequences and the PE will be treated as the inputs of the Transformer encoder. Generated from the encoder, the content features $ f_c $ and style features $ f_s $ then will be merged in the transfer module which is based on a Transformer decoder. Finally, the outputs can be obtained by decoding the fused features $ f_{cs} $ in a CNN-based decoder. In addition, to calculate the edge loss during the training step, the edge maps of the content images and the stylized images need to be extracted by a fine-designed edge extractor which will be introduced later.

In this part, we plan to present the overall architecture of the proposed STT first in Section \ref{section: Overall Architecture} and in Section \ref{section: Edge Extractor} introduce the edge extractor which is used to calculate the edge loss. Finally, the optimization strategy will be discussed in Section \ref{section: Network Optimization}.

\subsection{Overall Architecture}\label{section: Overall Architecture}
\begin{figure}[htbp]
\centerline{\includegraphics[width=1.0\linewidth]{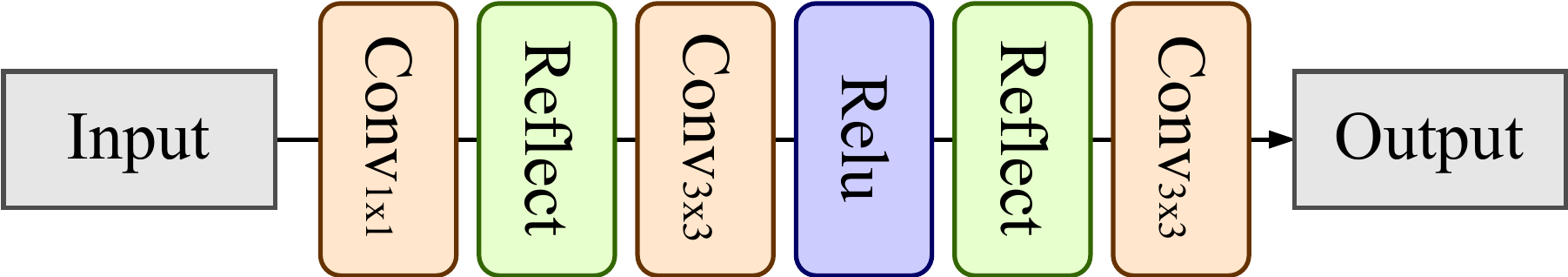}}
\caption{The illustration of Conv PE.}
\label{fig3}
\end{figure}

{\flushleft \bf Encoder}.
Before the process of the encoder, a tiny CNN-based module named Conv PE is applied to the content images and style images to extract the content-aware positional encoding. As depicted in Fig.~\ref{fig3}, Conv PE is composed by three convolutional layers, two reflections (padding) layers, and one ReLU activation layer. The main role of the reflection layers is to ensure that the size of results is consistent before and after processing while the convolutional layers are used to extract the content-aware positional encoding. As shown in Fig.~\ref{fig6}, we find that the results from the model with Conv PE are obviously better than that of the one without.

Different from the design of StyTr2 \cite{p45} which has two independent domain-specific encoders for the content images and style images, STT treats them as normal pictures with content \& style features and encodes them in one Transformer-based encoder. Given the input image in the shape of $ H \times W \times 3 $, the input will first be split into patches by the patch partition layer and then embedded into sequences linearly with the shape of $ \frac{HW}{8\times8} \times C $ (768 is the default value of $ C $). Adding the positional encoding from Conv PE, the resulting sequences will be fed to a three-layer Transformer encoder. The computation process of each layer can be defined as:
\begin{align}
    \hat{c}^{l} & = MSA(LN(c^{l-1})) + c^{l-1} \label{eq1} \\
    {c}^{l} & = MLP(LN(\hat{c}^{l})) + \hat{c}^{l} \label{eq2}    
\end{align}
where $ \hat{c}^{l} $ and $ {c}^{l} $ denote the outputs of MSA and MLP for layer $ l $ respectively; MSA represents the module of multi-head self-attention while MLP denotes the module of multi-layer perceptron; and $ LN $ means LayerNorm. After the three layers of the encoder, we obtain the content features $ f_c $ and style features $ f_s $ with the shape consistent before and after processing. 

{\flushleft \bf Decoder}.
Instead of upsampling  the stylized features $ f_{cs} $ to the original size in a single projection as \cite{p90} once did, STT follows \cite{p21,p37,p41} to utilize a mirrored VGG to decode the stylized features gradually. Before the step of decoding, the stylized features need to be reshaped first for the sequence-like shape $ \frac{HW}{8\times8} \times C $. After the three stages of upsampling and refining, we obtain the stylized images $ I_{cs} $ with the shape of $ H \times W \times 3 $.

{\flushleft \bf Transfer Module}.
The transfer module is used to merge the content features $ f_c $ and style features $ f_s $. We introduce the transfer module of S2WAT \cite{p89} as the means to fuse the features. As depicted in Fig.~\ref{fig2} (b), the transfer module consists of three layers of the Transformer decoder layers, and each layer is mainly composed by an MSA module, an MHA module, and an MLP module. The computational process can be defined as: 
\begin{gather}
    \hat{x}^{l} = MSA(LN(x^{l-1})) + x^{l-1} \notag \\
	Q = LN(\hat{x}^{l}) \cdot W_{Q} \notag \\
	K,V = y \cdot W_{K},\,\,  y \cdot W_{V} \notag \\
	\tilde{x}^{l} = MHA(Q,K,V) + \hat{x}^{l} \notag \\
	{x}^{l} = MLP(LN(\tilde{x}^{l})) + \tilde{x}^{l} \label{eq3}    
\end{gather}
where $ \hat{x}^{l} $, $ \tilde{x}^{l} $, and $ {x}^{l} $ represent the results of MSA, MHA, and MLP for layer $ l $, respectively; $ y $ denotes the style features; $ W_{Q} $, $ W_{K} $, and $ W_{V} $ are the projection matrices for $ Q $, $ K $, and $ V $; $ Q $, $ K $, and $ V $ denote the query, key, and value vectors. Leveraging the fusion effects of the cross attention, the stylized features $ f_{cs} $ can be received.

\subsection{Edge Extractor}\label{section: Edge Extractor}
To make the content structure of the stylized images clear, we design a novel edge loss to enhance the edge of the objects in output images. Before calculating the edge loss, the edge maps suitable for style transfer need to be captured by a fine-designed edge extractor. Different from the tasks like edge detection or contour extraction, the content details of the outputs in image style transfer are probably not the same as that of the content images, especially the background which may has the artistic patterns from the style images. The results will be blurred if we take the similarity between the edge maps from the content images and the stylized images as the optimization target directly. Therefore one of the problems that need to be solved is to filter out the place where the main structure of the content images does not exist. A mask operation is introduced to cope with this problem. As shown in \eqref{eq6}, all of the edges in the edge maps of the stylized images ($ edg^{\prime}\mbox{-}I_{cs} $) that are not exist in the corresponding place of the edge maps of content images ($ edg\mbox{-}I_c $) will be masked out. Furthermore, we also set a threshold to exclude the weak responses of edge maps which may play a role as noise. The overall computational process can be defined as:
\begin{gather}
    edg\mbox{-}I_c = threshold(lap(I_c), \, \tau) \label{eq4} \\
    edg^{\prime}\mbox{-}I_{cs} = threshold(lap(I_{cs}),  \, \tau) \label{eq5} \\
    edg\mbox{-}I_{cs} = mask(edg^{\prime}\mbox{-}I_{cs}, \, edg\mbox{-}I_c) \label{eq6}
\end{gather}
where $ edg\mbox{-}I_c $ and $ edg\mbox{-}I_{cs} $ are the edge maps of the content and stylized images respectively; $ lap $ denotes the Laplacian operator and $ threshold $ represents the function that sets 0 to the responses where the value is smaller than the threshold parameter $ \tau $ (0.2 is set as the default value). After the above steps, we obtain the refined edge maps to be used in calculating the edge loss.

\subsection{Network Optimization}\label{section: Network Optimization}
The main purpose of image style transfer is to maintain the structure of the content images while transferring the artistic patterns to the stylized results from the style images. To achieve this target, we follow \cite{p21} to construct two perceptual losses to measure the content differences between the stylized images and the content images as well as the style differences between the stylized images and the style images. Furthermore, we also adopt the identity losses \cite{p37} to enrich the content details and style patterns of the stylized images. Finally, the proposed edge loss is equipped to enhance the content structure further. As shown in \eqref{eq7}, the whole loss function can be defined as: 
\begin{align}
    \mathcal{L}_{total} = &\lambda_{c}\mathcal{L}_{content} + \lambda_{s} \mathcal{L}_{style} + \notag \\
    &\lambda_{id1}\mathcal{L}_{id1} + \lambda_{id2} \mathcal{L}_{id2} + \lambda_{edg}\mathcal{L}_{edg} \label{eq7}
\end{align}
where the $ \lambda_{c} $, $ \lambda_{s} $, $ \lambda_{id1} $, $ \lambda_{id2} $, and $ \lambda_{edg} $ are the weights of losses; $ \mathcal{L}_{content} $ and $ \mathcal{L}_{style} $ denote the perceptual losses; $ \mathcal{L}_{id1} $ and $ \mathcal{L}_{id2} $ are the identity losses; $ \mathcal{L}_{edg} $ represents the edge loss and we only apply the edge loss in the situation when the results are apparently blurred. We set $ \lambda_{c} $, $ \lambda_{s} $, $ \lambda_{id1} $, $ \lambda_{id2} $, and $ \lambda_{edg} $ to 1, 3, 50, 1, and 5000 to alleviate the impact of magnitude differences.

{\flushleft \bf Perceptual Loss}.
Similar to \cite{p21}, we leverage a pretrained VGG19 to extract the feature maps of the content and style images which are used to calculate the perceptual losses. In our model, the layer $ Relu\_4\_1 $ and $ Relu\_5\_1 $ are used to calculate the content perceptual loss while the layer $ Relu\_1\_1 $, $ Relu\_2\_1 $, $ Relu\_3\_1 $, $ Relu\_4\_1 $, and $ Relu\_5\_1 $ are used to calculate the style perceptual loss. One thing that needs to be attended to is that the mean-variance channel-wise normalization is applied on the feature maps before the calculation of the content perceptual loss. The perceptual losses can be defined as: 
\begin{align}
    \mathcal{L}_{content} = \sum_{l \in C} &\Vert \overline{\phi_{l}(I_{cs})} - \overline{\phi_{l}(I_{c})} \Vert_{2}^{} \label{eq8} \\[4px]
    \mathcal{L}_{style} = \sum_{l \in L} &\Vert \mu(\phi_{l}(I_{cs})) - \mu(\phi_{l}(I_{s})) \Vert_{2} + \notag \\
    &\Vert \sigma(\phi_{l}(I_{cs})) - \sigma(\phi_{l}(I_{s})) \Vert_{2} \label{eq9}
\end{align}
where the $ C $ and $ L $ are the layers of the pretrained VGG which are concerned to calculate the content and style perceptual losses respectively; $ \phi_{l} $ denotes the feature maps of the $l$-th layer in the pretrained VGG; $ \mu $ and $ \sigma $ are the mean and variance of the features; and the overline represents the mean-variance channel-wise normalization.

{\flushleft \bf Identity Loss}.
Following the work of \cite{p37}, a pair of identity losses are constructed to learn the relationship between the content and style representations. The identity losses are defined as :
\begin{gather}
    \mathcal{L}_{id1} = \Vert I_{cc} - I_{c}\Vert_{2}^{} + \Vert I_{ss} - I_{s}\Vert_{2}^{} \label{eq10} \\[4px]
    \mathcal{L}_{id2} = \sum_{l \in L} \Vert \phi_{l}(I_{cc}) - \phi_{l}(I_{c})\Vert_{2}^{} + \Vert \phi_{l}(I_{ss}) - \phi_{l}(I_{s})\Vert_{2}^{} \label{eq11}
\end{gather}
where $ I_{cc} $ ($ I_{ss} $) denotes the stylized images from a common pair of content (style) images. Specifically, the original content (style) image is expected when we feed two of the same content (style) images to the model. As shown in \eqref{eq11}, this operation is also applied on feature maps from the pretrained VGG. And the layer $ Relu\_1\_1 $, $ Relu\_2\_1 $, $ Relu\_3\_1 $, $ Relu\_4\_1 $, and $ Relu\_5\_1 $ are used to calculate the second identity loss.

{\flushleft \bf Edge Loss}.
To enhance the edge of the objects when the original results from STT are obviously blurred, we design an edge loss to cope with this problem. As depicted in Section \ref{section: Edge Extractor}, the edge maps are computed by the Laplacian operator first and then refined by a threshold function and a mask operation successively. After we obtain the refined edge maps, the edge loss can be computed in the following process:
\begin{gather}
    \mathcal{L}_{edg} = \Vert edg\mbox{-}I_c - edg\mbox{-}I_{cs} \Vert_{2}^{}
\end{gather}
where $ edg\mbox{-}I_c $ and $ edg\mbox{-}I_{cs} $ are the refined edge maps of the content and stylized images respectively. As shown in Fig.~\ref{fig1} and Fig.~\ref{fig7} (columns 3 and 6), applying the edge loss on STT can obviously improve the edges of blurred results.

\section{Experiments}
\subsection{Implementation Details}
{\flushleft \bf Datasets}.
MS-COCO \cite{p84} is used as the content dataset while WikiArt \cite{p85} is used as the style dataset. We randomly select 80000 images of each dataset to build the training datasets. During the process of training, the input image will be resized to 512 on the shorter side first and then randomly cropped into $ 224 \times 224 $. While in the process of testing, inputs of any size are accepted.

{\flushleft \bf Training Information}.
Pytorch framework is used to implement STT and 40000 iterations are taken to complete the training. With a batch size of 4  and an initial learning rate of 1e-4, we use an Adam optimizer \cite{p87} to train the network and the warmup strategy \cite{p88} to adjust the learning rate. The training step is taken about 10 hours on a single Tesla V100 GPU. We also calculate the reference time (see the last row of Table \ref{table:1}) of different image style transfer models with one Tesla P100 GPU.

\subsection{Style Transfer Results}
In order to demonstrate the style transfer effect of the proposed STT, we make a comparison between the results from the proposed STT and the state-of-the-art arbitrary style transfer methods, including AdaIn \cite{p21}, WCT \cite{p22}, SANet \cite{p37}, MCC \cite{p41}, ArtFlow \cite{p43}, IEST \cite{p42}, CAST \cite{p46}, StyTr2 \cite{p45}, and S2WAT \cite{p89} .

\begin{figure*}[htbp]
\centerline{\includegraphics[width=1.0\linewidth]{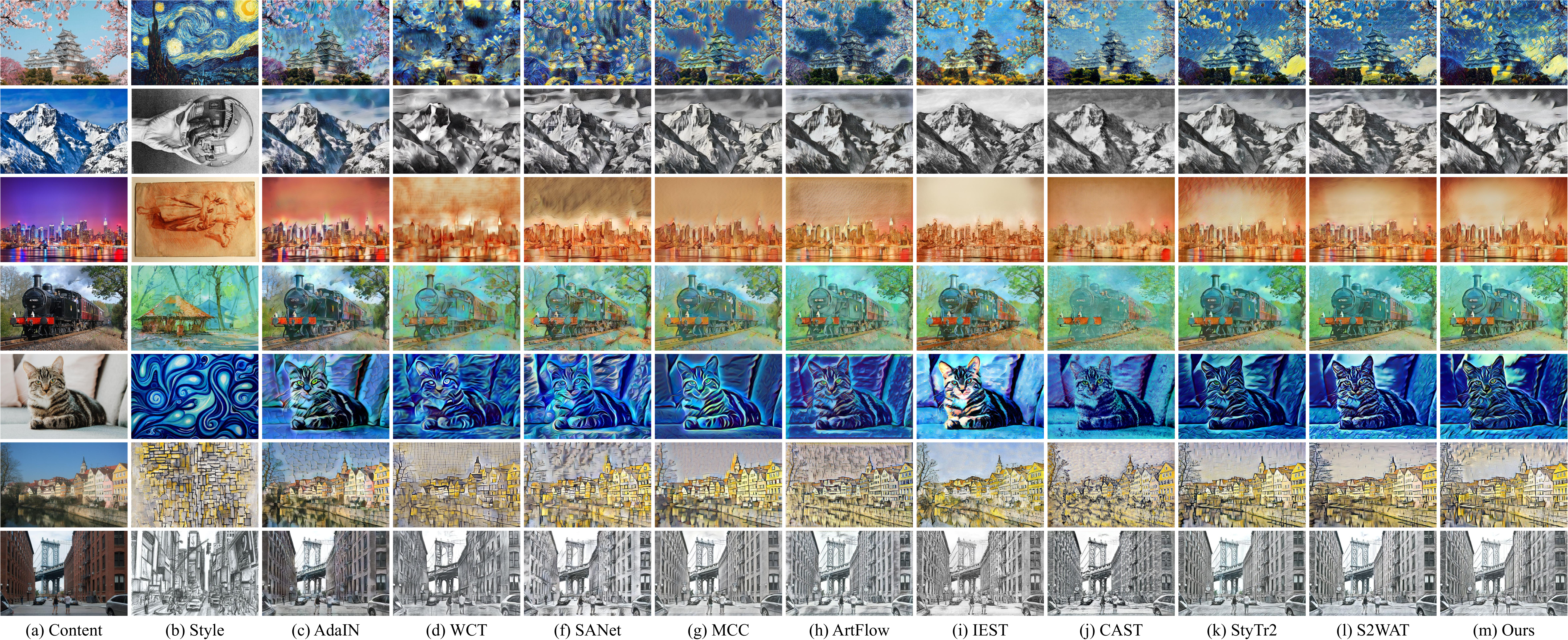}}
\caption{The visual comparison of the state-of-the-art arbitrary style transfer algorithms.}
\label{fig4}
\end{figure*}

{\flushleft \bf Qualitative Comparison}.
The results of the qualitative comparison are presented in Fig.~\ref{fig4}. Although the different methods fulfill the image style transfer in different ways, they all achieve colorful results. Due to the over-simplified alignment of the second-order statistics, AdaIN can not draw sufficient style patterns on the content images. By applying the alignment process on the style feature space with whitening and coloring operations, WCT attracts more artistic characteristics but damages the content details. Inspired by the attention mechanism, SANet transfers adequate style features to the content images but the structure is not ideal sometimes. MCC suffers from an overflow issue for the lack of linear operations. In conjunction with the projection flow network, ArtFlow is capable of producing content-unbiased results but sometimes may generate undesired patterns on the borders. Different from other methods which train the models with perceptual losses or identity losses, IEST and CAST adopt the contrastive learning strategy and make favorable effects sometimes. But in some cases, the results fail to obtain plentiful style representations. Transformer-based methods find a better balance between content and style. With the Transformer-based encoder and transfer module, StyTr2 and S2WAT both achieve satisfying effects while S2WAT may lose some style patterns and StyTr2 drops content details in some places. As shown in the last column of Fig.~\ref{fig4}, STT preserves the fine content details while sufficient artistic characteristics are transferred.

{\flushleft \bf Quantitative Comparison}.
In this part, the content differences between the stylized images and the content images are computed as an indirect metric to measure the content quality while the style differences between the results and the style images are calculated as an implicit metric to evaluate the style quality. The identity losses are also taken into consideration playing a role as the auxiliary metrics to judge the ability to preserve content/style features. As shown in Table \ref{table:1}, S2WAT achieves the lowest content loss while STT and SANet outperform the other methods on style quality. Compared with the CNN-based models, the Transformer-based methods have obvious advantages in identity losses. Due to the ability of completely reversible transformation, ArtFlow does not use identity losses. Although ArtFlow can produce content-unbiased results, STT outperforms it on style quality. In summary, STT can preserve both the content details from the content image as well as the style patterns from the style images.

\begin{table*}[htbp]
    \begin{center}
         \resizebox{0.8\linewidth}{!}{
         \begin{tabular}{c c c c c c c c c c c c}
            \hline
            Method & Ours & S2WAT & StyTr2 & CAST & IEST & ArtFlow & MCC & SANet & WCT & AdaIN \\
            \hline
            \textit{Content Loss} $\downarrow$ & 2.18 & \pmb{1.66} & 1.83 & 2.07 & 1.81 & 1.93 & 1.92 & 2.16 & 2.56 & \underline{1.71}  \\
            \textit{Style Loss} $\downarrow$ & \underline{1.35} & 1.74 & 1.52 & 4.33 & 2.72 & 1.90 & 1.70 & \pmb{1.11} & 2.23  & 3.50 \\
            \textit{Identity Loss 1} $\downarrow$  & \pmb{0.16} & \pmb{0.16} & \underline{0.26} & 1.94 & 0.91 & 0.00 & 1.07 & 0.81 & 3.01  & 2.54 \\
            \textit{Identity Loss 2} $\downarrow$ & \underline{1.55} & \pmb{1.38} & 3.10 & 18.72 & 7.16 & 0.00 & 7.72 & 6.03 & 21.88 & 17.97 \\
            \textit{Time(seconds)} $\downarrow$ & 0.270 & 0.558 & 0.237 &  \pmb{0.042} & \underline{0.061} & 0.325 & 0.078 & \underline{0.061} & 0.590 & \pmb{0.042} \\
            \hline
         \end{tabular}
         }
         \vspace{-4mm}
    \end{center}
    \caption{Quantitative comparison between the results from different image style transfer methods. The loss values above are all computed on 400 random samples average and the reference time is calculated on a hundred random samples in a resolution of $ 512 \times 512 $. The bold font marks the best values while the underline shows the second-best values.}
    \label{table:1}
\end{table*}

\subsection{Content Leak}
After repeated stylization with the same pair of content and style images, CNN-based methods will suffer the problem of content leak that the content structure will drop gradually as the number of experimental rounds grows. An et al. \cite{p43} utilize the projection flow network, a kind of network which is able to achieve completely reversible transformation, to settle the content leak problem. However, strict reversibility may have an undesired impact on the stylizing process. With the ability to capture long-range dependencies, StyTr2 \cite{p45} and S2WAT \cite{p89} are demonstrated to be capable of alleviating the content leak problem.

\begin{figure*}[htbp]
\centerline{\includegraphics[width=1.0\linewidth]{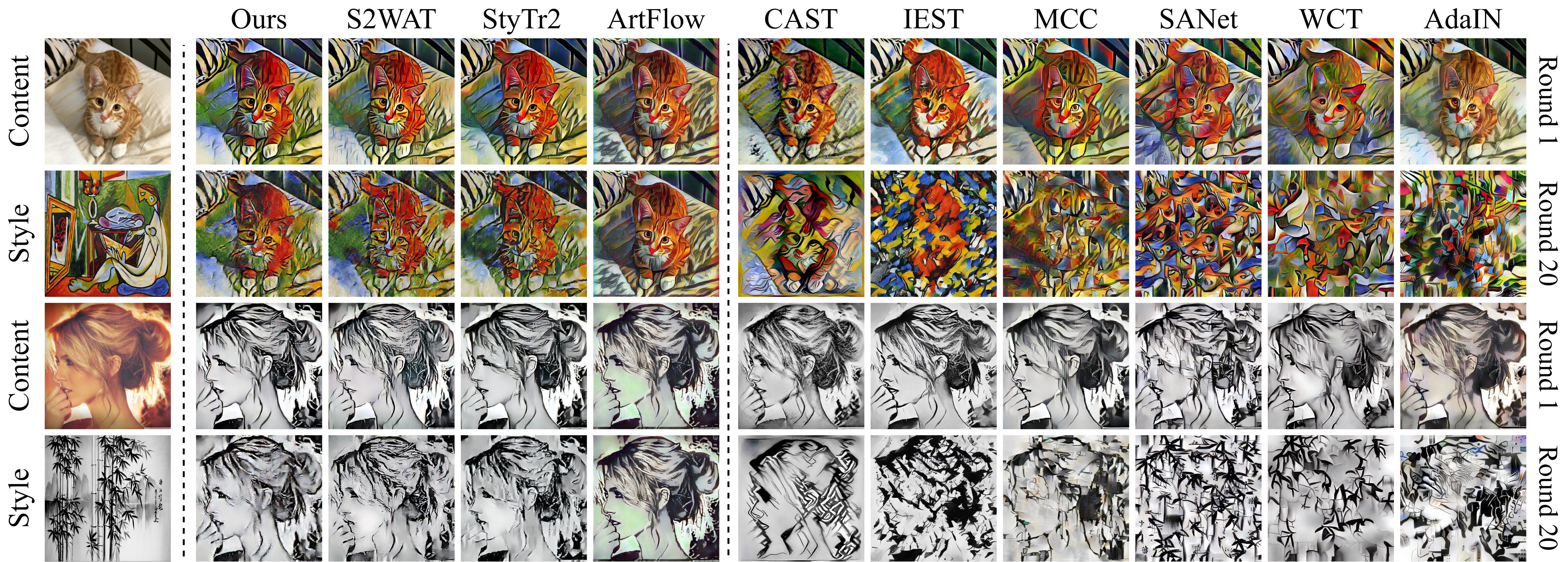}}
\caption{Visualization of the content leak problem.}
\label{fig5}
\end{figure*}

To examine the stylizing effects on the content leak issue, we make a comparison with the CNN-based method \cite{p21,p22,p37,p41,p42,p46}, the Flow-based method \cite{p43}, and the Transformer-based methods \cite{p45,p89}. As depicted in Fig.~\ref{fig5}, the results from the 1st and the 20th rounds of repeated stylization have been presented. All the methods can keep the content details well after the 1st stylizing process except that the results from AdaIN and ArtFlow are to some degree lack of style features. However, after the 20th round of the stylizing process, the CNN-based methods fail to preserve the content structure and the results are apparently blurred. Compared to the completely content-unbiased ArtFlow, the Transformer-based StyTr2, S2WAT, and the proposed STT still drop the content details slightly but the results are obviously superior to that of the CNN-based methods. Therefore, the proposed STT can preserve both the content structure and the style features while capable of alleviating  the content leak problem.

\subsection{Ablation Study}
\begin{figure}[t]
\centerline{\includegraphics[width=1.0\linewidth]{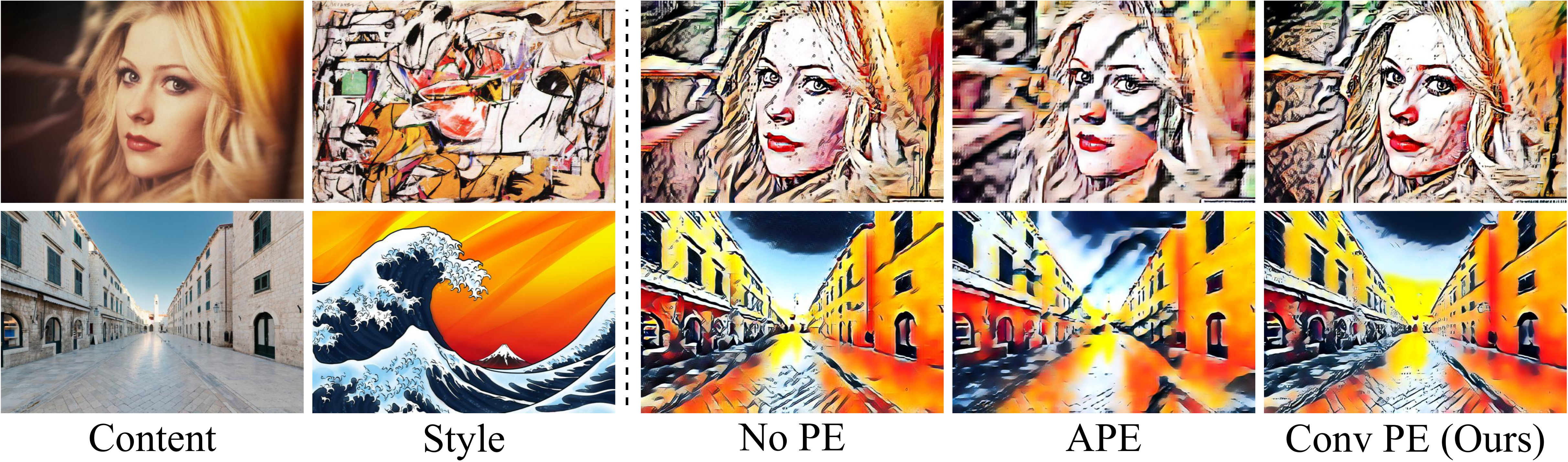}}
\caption{Comparison between the results from different types of PE.}
\label{fig6}
\end{figure}

{\flushleft \bf Conv PE}.
Positional encodings (PE) are important for Transformer-based models, which provide information on locations. There are two types of absolute positional encoding (APE) that are widely used: functional \cite{p48} and parametric \cite{p50} positional encoding. As Deng et al. \cite{p45} have discussed in StyTr2, the functional APE, such as the sinusoidal APE, will result in vertical track artifacts due to the large positional deviation. And we examine the parametric APE whose results are shown in Fig.~\ref{fig6} (column 4). Some undesired patterns that do not vary substantially with the inputs appear on the outputs. Due to the unsatisfactory performance of the functional APE and parametric APE, we propose a positional encoding based on convolutional operations (Conv PE), and the results are presented in Fig.~\ref{fig6} (column 5). Because the CAPE needs to work with the transfer module while the transfer module of STT does not have the interface of PE, we do not conduct the experiments on CAPE.

As depicted in Fig.~\ref{fig6}, the strokes of the results from the model without PE are obviously thicker than that from the model with Conv PE. Furthermore, there are a few vertical track artifacts on the edge of objects in images (row 1 column 3). For the results from the model with parametric APE, the background is blurry and a sort of undesired pattern makes the pictures unsightly. By contrast, the results from STT fix these problems and preserve both the content details and style features.

{\flushleft \bf Edge Loss}.
When the results of image style transfer are blurred, applying the edge loss on STT can improve picture clarity obviously. As depicted in Fig.~\ref{fig1}, the model without the edge loss erases the majority of content details in the content images, such as the windows on buildings (row 1 column 3) and the letters on the billboards (row 2 column 3). In contrast, these details are well preserved when the edge loss is equipped (column 4).

\begin{figure*}[htbp]
\centerline{\includegraphics[width=0.9\linewidth]{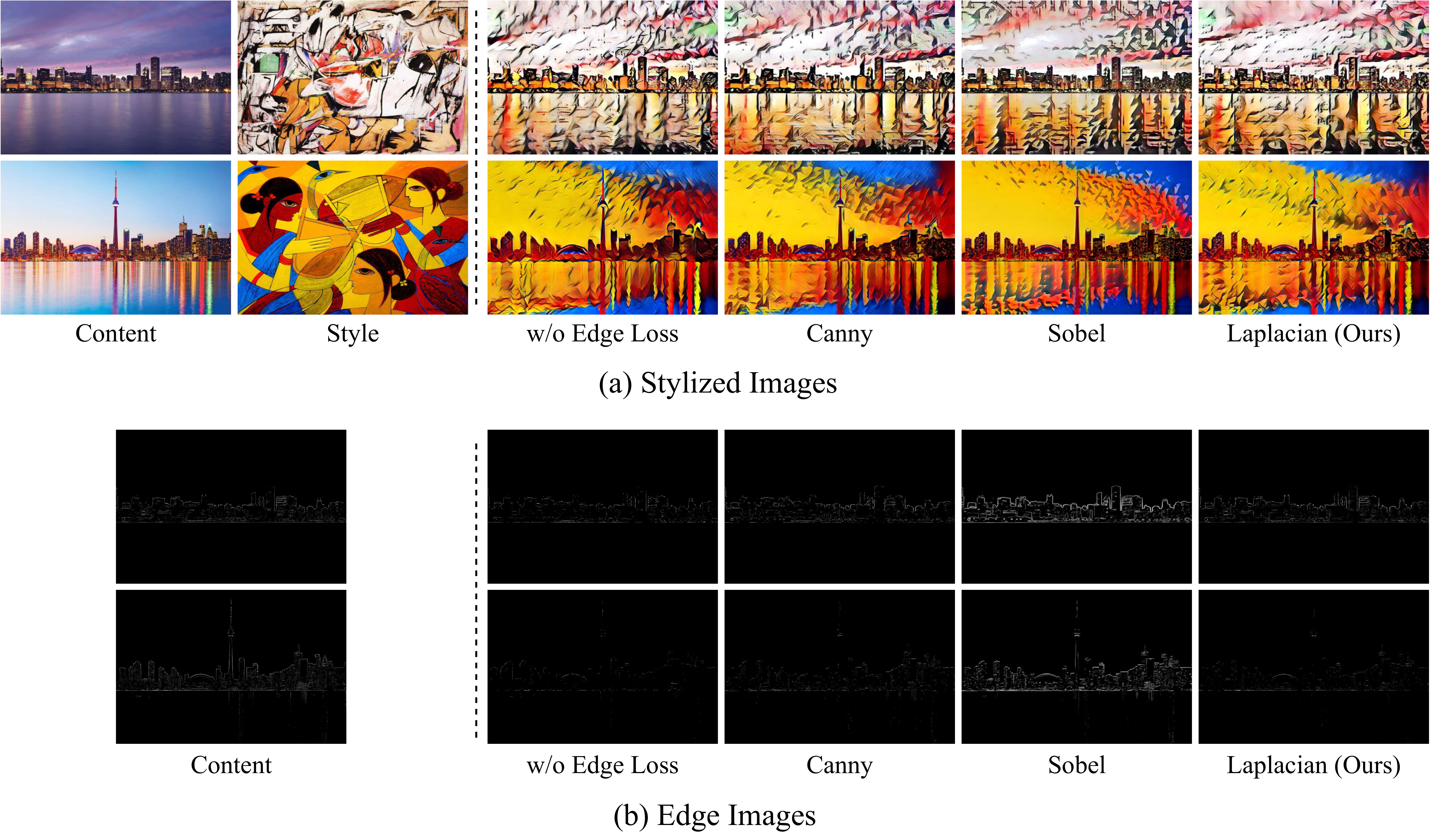}}
\caption{Comparison between the results using different edge detection operators.}
\label{fig7}
\end{figure*}

Besides the comparison between the models with and without the edge loss, we also compare the operators to extract the edge maps which are the important step to form the edge loss. As depicted in Fig.~\ref{fig7} (a), the operator of Canny, Sobel, and Laplacian are taken into consideration. A kind of hollow stroke appears on the results based on the Canny operator (see column 4) while the results on the Laplacian operator can produce natural and fine strokes. The clearest result though the model based on the Sobel operator can generate, unpleasant patterns, such as the vertical/horizontal tracks and the blurred strokes, appears in the stylized images (see column 5). For the outputs based on the Laplacian operator which is applied to the edge loss, the strokes are natural and the structure of objects is clear which demonstrate the performance of the edge loss.

In addition, we also provide the edge maps calculated by the edge extractor where a phenomenon can be easily found that the edges of the results from the model applying the edge loss will be much richer than that of the results from models without the edge loss.

\section{Conclusion}
In this work, we proposed a Transformer-based method named STT for arbitrary image style transfer. The proposed STT has a Transformer-based encoder that can encode both the content and style images capturing the long-range information between them. A content-aware positional encoding scheme (Conv PE) based on the convolutional operations is applied to the encoder to provide the positional information. To overcome the problem that the results of image style transfer are blurred in some cases, a novel edge loss is presented to improve the clarity of the stylized images. As another new method based on Transformer, STT is capable of producing vivid stylized images with fine content details and sufficient style features while alleviating the content leak problem.

{\small
\bibliographystyle{ieee_fullname}
\bibliography{ref}
}

\end{document}